\documentstyle[11pt]{article}

\begin{document}

\title{Estimating Performance of Pipelined \\
       Spoken Language Translation Systems}

\author{
Manny Rayner and David Carter\\
SRI International\\
Suite 23, Millers Yard\\
Cambridge CB2 1RQ, UK\\
\verb!\{manny,dmc\}@cam.sri.com!
\and
Patti Price\\
SRI International\\
333 Ravenswood Ave.\\
Menlo Park, CA 94025, USA\\
\verb!pprice@speech.sri.com!
\and
Bertil Lyberg\\
Telia Research AB\\
Rudsj\"oterassen 2\\
S-136 80 Haninge\\
Sweden\\
\verb!Bertil.Lyberg@haninge.trab.se!}

\maketitle

\begin{abstract}
Most spoken language translation systems developed to date rely on a
pipelined architecture, in which the main stages are speech
recognition, linguistic analysis, transfer, generation and speech
synthesis. When making projections of error rates for systems of this
kind, it is natural to assume that the error rates for the individual
components are independent, making the system accuracy the product of
the component accuracies.

The paper reports experiments carried out using the SRI-SICS-Telia
Research Spoken Language Translator and a 1000-utterance sample of
unseen data. The results suggest that the naive performance model
leads to serious overestimates of system error rates, since there are
in fact strong dependencies between the components. Predicting the
system error rate on the independence assumption by simple
multiplication resulted in a 16\% proportional overestimate for all
utterances, and a 19\% overestimate when only utterances of length 1-10
words were considered.
\end{abstract}

\section{INTRODUCTION}\label{Section:introduction}

Most spoken language translation systems rely on a pipelined
architecture, including speech recognition, linguistic analysis,
transfer, generation and speech synthesis.  A major advantage is that
components can be developed and tested independently.  This is
particularly important for spoken language translation, since
expertise in multiple languages is not often found in the same
location.  An obvious disadvantage is brittleness: if one model fails
to produce or pass on a correct interpretation, the whole translation
process fails.  To obtain modularity as well as robustness our system
consists of modules with multiple outputs and mechanisms for using
additional knowledge sources to reorder multiple inputs.  We have used
several statistical and other automatic methods to model knowledge
sources within the modules.

When making projections of error rates for systems of this kind, it is
natural to assume that the error rates for the individual components
are independent, making the system accuracy the product of
the component accuracies. Here, we will produce experimental evidence
suggesting that this simple model leads to serious overestimates of
system error rates, since there are in fact strong dependencies
between the components. For example, if an utterance fails
recognition then, had it been recognized, it would have had a higher
than average chance of failing linguistic analysis; similarly,
utterances which fail linguistic analysis due to incorrect choice in
the face of ambiguity are more likely to fail during the transfer and
generation phases if the correct choice is substituted. Intuitively,
utterances which are hard to hear are also hard to understand and
translate.

The experiments reported were carried out on the SRI-SICS-Telia
Research Spoken Language Translator
\cite{SLT-HLT-93,SLT-EUROSPEECH,SLT-first-year}, using a 1000-utterance
sample of previously unseen data. Processing was split into four
phases, and the partial results for each phase evaluated by skilled
judges. Where feasible (for example, for recognition), a correct
alternative was supplied when a processing phase produced an incorrect
result, and processing restarted from the alternative. This made it
possible to perform statistical analysis contrasting the results of
inputs corresponding to correct and incorrect upstream processing.

The results showed that dependencies, in some instances quite
striking, existed between the performances of most pairs of phases.
For example, the error rates for the linguistic analysis phase,
applied to correctly and incorrectly recognized utterances
respectively, differed by a factor of about 3.5; a chi-squared test
indicated that this was significant at the P=0.0005 level. The
dependencies existed at all utterance lengths, and were even stronger
when evaluation was limited to the portion of the corpus consisting of
utterances of length 1-10 words. Predicting the system error rate on
the independence assumption by simple multiplication resulted in a 16\%
proportional overestimate for all utterances, and a 19\% overestimate
for the 1-10 word utterances.

The rest of the paper is structured as follows.
Section~\ref{Section:SLT} gives a brief overview of the Spoken
Language Translator.  Section~\ref{Section:Experiments} presents a
detailed description of the experiments carried out, and
Section~\ref{Section:Summary} summarizes and concludes.

\section{THE SPOKEN LANGUAGE TRANSLATOR}\label{Section:SLT}

The Spoken Language Translator (SLT) is a pipelined speech-to-speech
translation system developed by SRI International, the Swedish
Institute of Computer Science, and Telia Research AB, Stockholm under
sponsorship from Swedish Telecom (Televerket N\"at); it translates
utterances in the air travel planning (ATIS) domain from spoken
English to spoken Swedish, using a vocabulary of about 1500 words.
Work on the project began in June 1992. The system is constructed from
a set of general-purpose speech and language processing components.
All the components existed prior to the start of the project; they
have been adapted to the ATIS speech translation task in ways
described at length elsewhere
\cite{SLT-HLT-93,SLT-first-year}. In most cases, the customization
process was fairly simple, and was performed using semi-automatic
training methods. The main components are the SRI DECIPHER(TM) system
(speech recognition); two copies of the SRI Core Language Engine (CLE)
(source and target language processing); and the Telia Research
Prophon system (speech synthesis).

The speech translation process begins with the SRI DECIPHER(TM)
system, based on hidden Markov modeling and a progressive search
\cite{Murveit:93,Digalakis:94}.  It outputs to the source
language processor an N-best list of sentence hypotheses generated
using acoustic and bigram language model scores. N is normally set to
a value between 5 and 10.

The source-language (English) copy of the CLE then performs linguistic
analysis on all the utterance hypotheses in the N-best list. The CLE
is a sophisticated unification-based language processing system which
incorporates a broad-coverage domain-independent grammar for English
\cite{CLE}. In the SLT system, the general CLE grammar is
specialized to the domain using the Explanation-Based Learning (EBL)
algorithm \cite{EBL-IJCAI-91}. The resulting grammar is parsed using
an LR parser \cite{Christer-thesis}, giving a decrease in analysis
time, compared to the normal CLE left-corner parser, of about a factor
of ten. The specialization process results in a small loss of
grammar coverage compared to the original grammar, the size of the
coverage loss being dependent on the size and nature of the training
corpus used.

After the linguistic analysis phase has been completed, each utterance
hypothesis is associated with a (possibly empty) set of semantic
analyses expressed in a predicate/argument style notation called Quasi
Logical form (QLF). The most plausible analysis (and hence,
implicitly, the most plausible utterance hypothesis) is then selected
by the ``preference module''. This module applies a variety of
preference functions to each analysis, and combines their scores
using scaling factors trained using a combination of least-squares
optimization and hill-climbing \cite{Alshawi+Carter:93,HLT-NBEST}. The
training material for both the Explanation-Based Learning
specialization process and the preference module comes from a
``treebank'' of about 5000 hand-verified examples.

The QLF selected by the preference module is passed to the transfer
component, which uses a set of non-deterministic unification-based
recursive rewriting rules to derive a set of possible corresponding
target-language (Swedish) QLFs \cite{BCI}. The preference component is
then called again to select the most plausible transferred QLF. This
is passed to a second copy of the CLE, loaded with a Swedish
grammar, to generate a target-language text string. The Swedish
grammar has been adapted fairly directly from the English one
\cite{Gamback+Rayner:92}. Generation is performed using the
Semantic Head-Driven algorithm \cite{SHD}, which simultaneously
constructs a phrase-structure tree as part of the generation process.
Finally, the output text string is passed to the Prophon speech
synthesizer \cite{PROPHON}, where it is converted into output speech
using a polyphone synthesis method. The phrase-structure tree is used
to improve the prosodic quality of the result.

The SLT system is described in detail in \cite{SLT-first-year}.

\section{EXPERIMENTS}
\label{Section:Experiments}

Many researchers working in the field of automatic spoken language
understanding have made the informal observation that utterances hard
for one module in an integrated system have a greater than average
chance of being hard for other modules; this effect is sometimes
referred to as ``synergy''. Quantitative studies are hard to come by,
however, which motivated the experiments described here. The test
corpus used was the 1001-utterance set of ATIS data provided for the
December 1993 ARPA Spoken Language Systems evaluations. This corpus was
unseen data for the present purposes.

We focussed our investigations on four conceptual functionalities in
the system: speech recognition, source language analysis, grammar
specialization, and transfer-and-generation. This breakdown was
motivated partially by the expense and tedium of judging intermediate
results by hand; ideally, we would have preferred a more fine-grained
division, for example splitting transfer-and-generation into two
phases. The results seem however adequate to illustrate our basic
point. The error rate for each functionality was defined as follows:
\begin{description}
\item[Speech recognition]

Proportion of utterances for which the preferred N-best hypothesis is
not an acceptable variant of the transcribed utterance. ``Acceptable
variant'' was judged strictly: thus for example substitution of ``a''
by ``the'' or {\it vice versa} was normally judged unacceptable, but
``all the'' instead of ``all of the'' would normally be acceptable.

\item[Source language analysis]

Proportion of input utterance hypotheses that do not receive a
semantic analysis. This neglects the problem that some semantic
analyses are incorrect; other studies (\cite{SLT-first-year}, Appendix
A) indicate that
of sentences for which some analyses are produced, around 5 to 10\%
are assigned only incorrect analyses.

\item[Grammar specialization]

Proportion of input utterance hypotheses receiving an analysis with
the normal grammar that receive no analysis with the specialized
grammar.

\item[Transfer-and-generation]

Proportion of input utterance hypotheses receiving an analysis with
the normal grammar that do not produce an acceptable translation.

\end{description}

The basic method for establishing correlations among processing
functionalities was to contrast results between two sets of inputs,
corresponding to i) correct upstream processing and ii) incorrect but
correctable upstream processing respectively. In the second case, the
input was substituted by input in which the upstream errors had been
corrected. The expectation was that in cases where an upstream error
had occurred the chance of failure in a given component would be
higher even if the upstream error were corrected; this indeed proved
to be the case.

The simplest example is provided by the linguistic processing phase.
Of the 1001 utterances, 789 were recognized acceptably, and 212
unacceptably. 706 of the utterance in the first group received a QLF
(89.5\%); when the 212 misrecognized utterances were replaced by the
correctly transcribed reference versions, only 135 (63.7\%) received a
QLF. Thus one can conclude that utterances failing recognition would
anyway be 3.5 times as likely to fail linguistic processing as well.
According to a standard chi-squared test, this result is significant
at the P=0.0005 level.

Moving on to the grammar specialization phase, there are two possible
types of upstream error for a given utterance: recognition can
fail, or the utterance can be out of coverage for the general
(unspecialized) grammar. Only the first type of error is correctable.
So the meaningful population of examples is the set of 706 + 135 = 841
utterances for which a QLF is produced assuming correct recognition.
Of the 706 correctly recognized examples, 653 (92.5\%) still produced
a QLF when the specialized grammar was used instead of the general
one. Of the 135 incorrectly recognized example, only 101 (74.8\%)
passed grammar specialization. The ratio of error rates, 3.4, is
similar to the one for linguistic analysis, and is also significant
at the P=0.0005 level.

For the transfer-and-generation phase, the population of meaningful
examples is again 841, but this time there are two types of correctable
upstream error: either recognition or grammar specialization can fail.
Of the 653 examples with no upstream error, 539 (82.5\%) produced
a good translation; of the 841 - 653 = 188 examples with a correctable
upstream error, 119 (63.3\%) produce a good translation. The ratio of
error rates, 2.1, is lower than for the linguistic analysis and
grammar specialization phases, but is still significant at the P=0.0005
level.

If we calculate error rates for each phase over the whole population
of meaningful examples (correct upstream processing + correctable
upstream errors), we get the following figures.
\begin{description}
\item[Recognition] 1001 examples; 789 successes; error rate = 21.2\%.
\item[Linguistic analysis] 1001 examples; 706 + 135 = 841 successes;
error rate = 15.9\%.
\item[Grammar specialization] 841 examples; 653 + 101 = 754 successes;
error rate = 10.3\%.
\item[Transfer and generation] 841 examples; 539 + 119 successes;
error rate = 21.8\%.
\end{description}
On the naive model, the error rate for the whole system should be
(1 - (1 - 0.212)(1 - 0.159)(1 - 0.103)(1 - 0.218)) = 0.535. In actual
fact, however, the error rate is (1 - 539/1001) = 0.462. Thus the
naive model overestimates the error rate by a factor of 0.535/0.462
= 1.16.

It is not immediately clear why these strong correlations exist. One
likely hypothesis which we felt needed investigation is that they are
a simple consequence of the known fact that accuracy in general
correlates strongly with utterance length, with long utterances
being difficult for all processing stages. If this were so, one would
expect the effect to be less pronounced if the long utterances were
removed. Interestingly, this does not turn out to be true. We
repeated the experiments using only utterances of 1 to 10 words in
length (688 utterances of the original 1001): the new results, in
summary, were as follows. All of them were significant at the P=0.0005
level.

\begin{description}
\item[Speech recognition]

577 utterances (83.9\%) were acceptably recognized.

\item[Linguistic analysis]

531 of the 577 acceptably recognized utterances (92.0\%) received a
QLF; 75 of the 111 unacceptably recognized utterances (67.6\%)
received a QLF. The ratio of error rates is 4.1.

\item[Grammar specialization]

497 of the 531 correctly recognized utterances receiving a QLF (93.6\%) passed
grammar specialization; 54 of the 75 relevant incorrectly recognized
utterances did so (72.0\%). The ratio of error rates is 4.4.

\item[Transfer and generation]

428 of the 497 utterances with no upstream error received a good
translation (86.1\%); 67 of the 109 utterances with a correctable
upstream error did so (61.5\%). The ratio of error rates is 2.8.

\end{description}
The naive model predicts a combined error rate of 45.1\%; the real
error rate is 37.8\%. Thus the naive model overestimates the error rate
by a factor of 1.19, an even larger difference than for the entire set.

A more plausible explanation for the correlations is that they arise
from the fact that all the components of the system are trained on,
and therefore biased towards, rather similar data. This training may
be automatic, or it may arise from system developers devoting their
efforts to more frequently occurring phenomena (a strategy followed
deliberately in adapting the Core Language Engine to the ATIS domain).
Even if training and test sentences formally outside the domain are
excluded from consideration, some sentences will still be more
``typical'' than others in that they employ more frequently occurring
words, word sequences, constructions and concepts. It is quite
probable that typicality at one level -- say, that of word N-grams,
making correct recognition more likely -- is strongly correlated with
typicality at others -- say, source language grammar coverage,
especially when specialized.

\section{SUMMARY AND CONCLUSIONS}
\label{Section:Summary}

There are several interesting conclusions to be drawn from the results
presented above. Most obviously, pipelined systems are clearly doing
rather better than the naive model predicts. More interestingly, the
experiments clearly show that the whole concept of evaluating
individual components of a pipelined system in isolation is more
complex than one at first imagines. Since all the components tend to
find the same utterances difficult, the upstream components act as a
filter which separate out the hard examples and pass on the easy ones.
Thus a test which measures the performance of a component in an ideal
situation, assuming no upstream errors, will in practice give a more
or less misleading picture of how it will behave in the context of the
full system. In general, downstream components will {\it always} have
a lower error rate than a test of this type suggests.

In particular, the performance of the language processing component of
a pipelined speech-understanding system is not something that can
meaningfully be measured in isolation. A clear understanding of this
fact allows development effort to be focussed more productively on
work that improves system performance as a whole.


\begin{thebibliography}{24}

\def\icassp{ {\it Proc. of the Inter. Conf. on
Acoust., Speech and Signal Proc.}}

\bibitem{SLT-first-year}
Agn\"as,~M-S., Alshawi,~H., Bretan,~I., Carter,~D.M.  Ceder,~K.,
Collins,~M., Crouch,~R., Digalakis,~V., Ekholm,~B., Gamb\"ack,~B.,
Kaja,~J., Karlgren,~J., Lyberg,~B., Price,~P., Pulman,~S., Rayner,~M.,
Samuelsson,~C. and Svensson,~T., {\it Spoken Language Translator:
First Year Report}, joint SRI/SICS technical report, 1994.

\bibitem{CLE} Alshawi,~H., {\it The Core Language Engine},
Cambridge, Massachusetts: The MIT Press, 1992.

\bibitem{Alshawi+Carter:93} Alshawi,~H. and Carter,~D.M.,
{\it Training and Scaling Preference Functions for Disambiguation}, To
appear in {\it Computational Linguistics}, 1995. Also available
as SRI technical report.

\bibitem{BCI}
Alshawi, H., Carter, D., Rayner, M. and Gamb\"ack, B.,
``Transfer through Quasi Logical Form'',
{\it Proc. 29th ACL}, Berkeley, 1991.

\bibitem{PROPHON} Ceder, K. and Lyberg, B.,
``Yet Another Rule Compiler for Text-to-Speech Conversion?'',
{\it Proc. ICSLP}, Banff, 1993.

\bibitem{Digalakis:94}
Digalakis,~V. and Murveit,~H.,
``Genones: Optimizing the Degree of Tying in a Large Vocabulary HMM
Speech Recognizer'', \icassp, 1994.

\bibitem{Gamback+Rayner:92}
Gamb\"ack, B. and Rayner, M.,
``The Swedish Core Language Engine'',
{\it Proc. 3rd NOTEX}, Link\"oping, 1992.

\bibitem{Murveit:93}
Murveit,~H., Butzberger,~J., Digalakis,~V. and Weintraub,~M.,
``Large Vocabulary Dictation using SRI's DECIPHER(TM)
Speech Recognition System: Progressive Search Techniques'',
\icassp, Minneapolis, Minnesota, April 1993.

\bibitem{SLT-HLT-93}
Rayner,~M., Alshawi,~H., Bretan,~I., Carter,~D.M.,
Digalakis,~V., Gamb\"ack,~B., Kaja,~J., Karlgren,~J.,
Lyberg,~B., Price,~P., Pulman,~S. and Samuelsson,~C.,
``A Speech to Speech Translation System Built From
Standard Components''.
{\it Proc.~ARPA workshop on Human Language Technology, 1993}

\bibitem{HLT-NBEST}
Rayner,~M., D.~Carter, V.~Digalakis and P.~Price, ``Combining
Knowledge Sources to Reorder N-Best Speech Hypothesis Lists''.
To appear in {\it Proc.~ARPA workshop on Human Language Technology, 1994}

\bibitem{SLT-EUROSPEECH}
Rayner,~M., Bretan,~I., Carter,~D., Collins,~M., Digalakis,~V.,
Gamb\"ack,~B., Kaja,~J., Karlgren,~J., Lyberg,~B., Price,~P.,
Pulman~S. and Samuelsson,~C., ``Spoken Language Translation with
Mid-90's Technology: A Case Study''. {\it Proceedings of Eurospeech
'93}, Berlin, 1993.

\bibitem{EBL-IJCAI-91} Samuelsson,~C. and Rayner,~M.,
``Quantitative Evaluation of Explanation-Based Learning as a
Tuning Tool for a Large-Scale Natural Language System''.
{\it Proc. 12th International Joint Conference on Artificial Intelligence}.
Sydney, Australia, 1991.

\bibitem{Christer-thesis}
Samuelsson,~C., {\it Fast Natural Language Parsing Using
Explanation-Based Learning}, PhD thesis, Royal Institute of
Technology, Stockholm, Sweden, 1994.

\bibitem{SHD} Shieber, S. M., van Noord, G., Pereira, F.C.N and Moore, R.C.,
``Semantic-Head-Driven Generation'',
{\it Computational Linguistics}, 16:30--43, 1990.

\end{thebibliography}
\end{document}